\documentclass{article}

\usepackage[preprint]{nips_2018}

\usepackage[utf8]{inputenc} 
\usepackage[T1]{fontenc}    
\usepackage{hyperref}       
\usepackage{url}            
\usepackage{booktabs}       
\usepackage{amsfonts}       
\usepackage{nicefrac}       
\usepackage{microtype}      

\usepackage{bbm}
\usepackage{algorithm}
\usepackage[noend]{algpseudocode}
\usepackage{mathtools}
\usepackage{graphicx}
\usepackage{subcaption} 
\usepackage{mwe}

\setcitestyle{square,numbers,comma}

\title{SARN: Relational Reasoning through Sequential Attention}

\author{
  Jinwon An  \\
  Department of Industrial Engineering\\
  Seoul National University\\
  Seoul, Republic of Korea\\
  \texttt{jinwon@dm.snu.ac.kr} \\
  \AND
   Sungwon Lyu\\
   Department of Industrial Engineering\\
  Seoul National University\\
  Seoul, Republic of Korea\\
   \texttt{lyusungwon@dm.snu.ac.kr} \\
  \And
  Sungzoon Cho\\
  Department of Industrial Engineering\\
  Seoul National University\\
  Seoul, Republic of Korea\\
  \texttt{zoon@snu.ac.kr}\\
}

\begin{document}

\maketitle

\begin{abstract}
This paper proposes an attention module augmented relational network called SARN(Sequential Attention Relational Network) that can carry out relational reasoning by extracting reference objects and making efficient pairing between objects. SARN greatly reduces the computational and memory requirements of the relational network by \citep{santoro2017simple}, which computes all object pairs. It also shows high accuracy on the Sort-of-CLEVR dataset compared to other models, especially on relational questions. 
\end{abstract}

\section{Introduction}
Relational reasoning is one of the fundamental building blocks for all kinds of human cognitive activity \citep{kemp2008discovery}. While representing relations and performing reasoning based on them is a challenging problem\citep{newell1980physical, harnad1990symbol}, many graph-based approaches tried to solve these problems \citep{scarselli2009graph, li2015gated}. These studies were primarily focused on solving problems using a sparse matrix representing relationships that form a network. Moreover, relational reasoning based on network structure requires clearly defined entities and relations, which is mostly not the case in the real world. 

We viewed relational reasoning as a series of decision processes. Consider a general setting where there are many objects in a scene and information on certain relationships has to be inferred based on a given question. The relational reasoning procedure begins with identifying the reference object based on the question. For example, given a certain image to answer the question "What is the closest object near to object A?", we first identify object A, the reference object, in the image. After the reference object is identified, it is compared with other objects to determine the relationship between each pair. Afterward, distance from object A to other objects will be calculated and sorted to find the nearest object. We do not need the relationship information between object B and object C to answer the question above. In other words, we focus our 'attention' only on the relevant relationships by filtering out other relationships that do not include the reference object. 


Previous research on relational neural networks that gave intuition to our study was \citep{santoro2017simple} and \citep{yang2016stacked}. \citep{santoro2017simple} proposed the relational module (RN) that tries to compute relational information by pairing each object representation with each other using a relational module. However, because all object pairs were put through the relational module, computational complexity is in $O(n^2)$ where $n$ is the number of objects. 

\citep{yang2016stacked} used sequentially stacked attention algorithms to focus on areas of images that are relevant to solving a given question. Attention maps enhance performance but also explains how the model views the image when reasoning. Although this incorporates the sequential reasoning process, it does not use an explicit relational reasoning module.

In this work, we propose an efficient relational reasoning algorithm that sequentially processes information using attention. The reference object is found using soft-attention, which is paired with other objects. Next, the relevant relationship between each object and the reference object is extracted. By only making object pairs that include the reference object, computational complexity is now in $O(n)$. Attention maps and relational module activation maps show that the results are much more interpretable, selectively showing high activation values in areas of the image that is relevant.

\section{Framework and formulation}
\label{framework and formulation}
We use the same notation as in \citep{santoro2017simple}: a convolutional neural network for pixel-wise object representation using feature maps (with coordinate vectors attached), $g_{\theta}$ for the relational module and $f_{\phi}$ for processing the aggregated relational information. An additional attention module that we propose for representing the reference object is $a_{\psi}$. When we use the word object, we refer to the pixel in the feature map, except for the reference object which is a weighted sum of the pixels.

First, we extract the reference object using soft-attention. The attention module takes the feature map and question embedding as input:
\begin{equation}
a_i = a_{\psi}(o_i, q) \quad i = 1, \cdots, n
\label{eq:attention}
\end{equation}
where $o_i$ is an object and $q$ is the question embedding. It outputs a softmax attention across the objects that locates the reference object. The reference object is represented as a weighted sum of the objects:

\begin{equation}
ro = \sum_{i=1, \cdots, n} a_i * o_i
\label{eq:reference_object}
\end{equation}

The reason why we use soft-attention instead of hard-attention is that each pixel of the feature map does not exactly correspond to one object. In other words, it is possible that the receptive field of a pixel in feature map does not contain an object entirely. It could be distributed among nearby pixels. Selecting only one pixel could force the object representation to be inconclusive. A weighted sum of soft-attention can adequately represent the reference object even in these situations. We check this idea in Section~\ref{sec:robust} by considering various image resolutions.

Next, We pair this reference object representation with other the objects by concatenating it channel-wise. As in $a_{\psi}$, the question embedding is concatenated to each pair and is fed to the relational module $g_{\theta}$:

\begin{equation}
g_{\theta output} = \sum_{i=1, \cdots, n} g_{\theta}(o_i, ro, q)
\label{eq:reference_object}
\end{equation}

\begin{figure}[t]
\includegraphics[width=\textwidth]{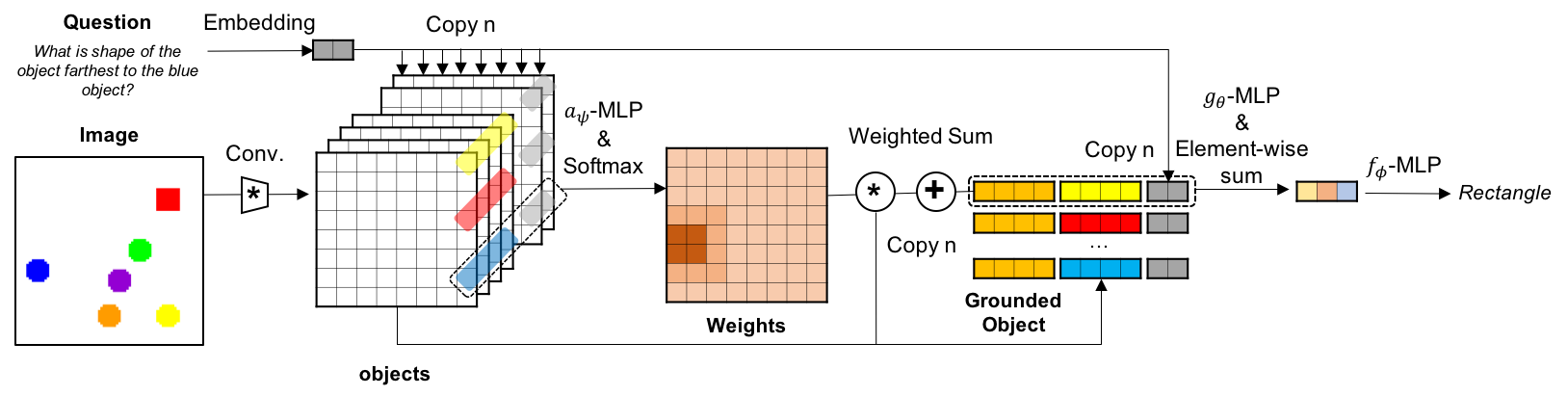}
  \caption{Model architecture overview}  
  \label{fig:model_overview}
  \centering  
\end{figure}

Figure~\ref{fig:model_overview} shows the overall model of our proposed model.

\section{Experiments}
\label{experiments}
\subsection{Dataset}
In our experiments, we used a dataset which is a modified version of Sort-of-CLEVR from \citep{santoro2017simple}. Each $75 \times 75$ image has 6 objects, whose shape is randomly assigned to be a square or a circle. 6 different colors were used to identify each object. Given a reference object identified by one of the 6 colors, 3 non-relational and 5 relational questions are generated. The non-relational questions are the same as in \citep{santoro2017simple}: (1) horizontal position, (2) vertical position, (3) shape. Relational questions of \citep{santoro2017simple} are (1) shape of the nearest object (2) shape of the furthest object (3) number of objects of the same shape. Additionally, (4) color of the nearest object, (5) color of the furthest object is also added to the relational question list. Sample questions are shown in Figure \label{fig:example_plots}. A total of 9800 images were generated for training and 200 for left for testing. Each image has 48 questions ($6 \times 3$ non-relational questions,$6 \times 5$ relational questions). Question vectors are represented by concatenating two one-hot encoding vectors, one for the color and the other for the question type.

\subsection{Models and parameters}{\tiny }
We ran three different models. The relational network of \citep{santoro2017simple}, a baseline model without object pairing, and our proposed model SARN. Our baseline model is different from that of \citep{santoro2017simple} which flattens out the CNN feature map and concatenate it with the question embedding. We used a different baseline model that takes individual objects as inputs for $g_{\theta}$ instead of paired inputs and is run through $f_{\phi}$.

The model parameters are the same for each model. CNN: 4 convolutional layers with 32 kernels, ReLU non-linearities, and layer normalization. $g_{\theta}$, $f_{\phi}$ and $a_{\psi}$: three-layer MLP with 128 hidden units per layer.

Test accuracy is shown in table Table ~\ref{tbl:acc_summary}. SARN shows higher accuracy in both non-relational and relational task. For detailed accuracy results for each type of question, see the Appendix.

\begin{table}[b]
\centering
\caption{Test accuracy}
\label{tbl:acc_summary}
\begin{tabular}{@{}llll@{}}
\toprule
model                                    & overall        & non-rel        & rel            \\ \midrule
SARN & \textbf{96.73} & \textbf{99.84} & \textbf{94.88} \\ 
RN                       & 93.56          & 99.81          & 89.83          \\ 
base line                                & 89.07          & 97.58          & 83.97          \\
\bottomrule
\end{tabular}
\label{tbl:accuracy_table}
\end{table}

\subsection{Reasoning inspection and interpretability}
Since our model runs in a sequential manner, we can examine the attention module and the relational module to verify whether reference objects are correctly retrieved, and important relationships are highlighted.

The attention map produced by $a_{\psi}$ is shown in Figure ~\ref{fig:example_plots}. It shows the weights of $a_{\psi}$. It correctly identifies the region of the reference object according to the question. We did not give only the color embedding vector for $a_{\psi}$ but used the whole concatenated question embedding vector. $a_{\psi}$ learned what information to use from the concatenated question embedding vector.

Inspecting the output of $g_{\theta}$ can show whether the most relevant pairs were identified regarding the question. To represent the average activation value for each object-reference pair, $g_{\theta}$ is summed up across channels. This shows the aggregated amount of activation values that each object-reference pair has produced.

\begin{figure*}[ht]
\centering
\begin{subfigure}[b]{\textwidth}
\centering   
\includegraphics[width=\textwidth]{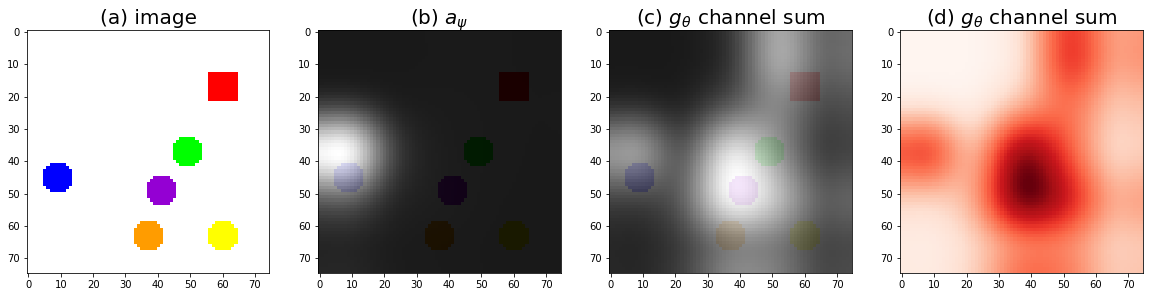}  
  \caption{What is shape of the object closest to the blue object?}  
  \label{fig:attention_blue_closest_shape}
\end{subfigure}

\begin{subfigure}[b]{\textwidth}
\centering   
\includegraphics[width=\textwidth]{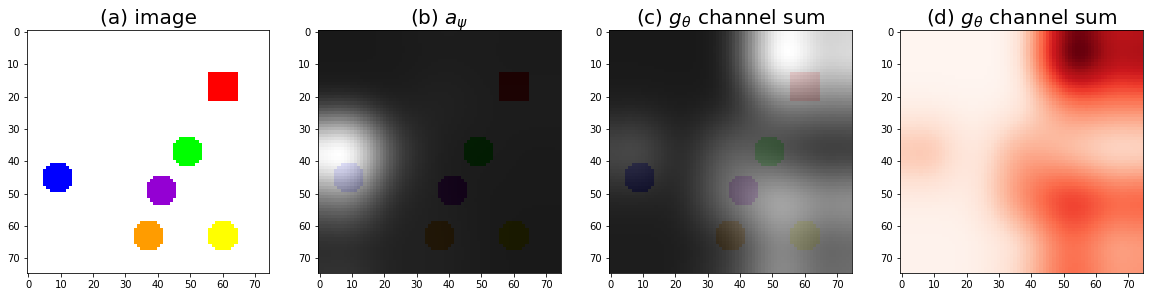} 
  \caption{What is shape of the object furthest to the blue object?}  
  \label{fig:attention_blue_furthest_shape}
\end{subfigure}
\caption{\textbf{Sample attention maps and relational module activation}: The first column shows the image. The second column shows the (upscaled) attention map of $a_{\psi}$ overlaid on the image. The third column shows $g_{\theta}$ summed up across channels overlaid on the image. The fourth column is used to emphasize and compare the amount of activation value of $g_{\theta}$ for each object}
\label{fig:example_plots}
\end{figure*}

Figure~\ref{fig:example_plots}~\subref{fig:attention_blue_closest_shape} shows that $a_{\psi}$ correctly picks the reference object as the blue object. Regarding channel sum value of $g_{\theta}$ , it also correctly exhibits high activation values for objects that are near the blue object. When the question is finding the furthest objects as in Figure~\ref{fig:example_plots}~\subref{fig:attention_blue_furthest_shape}, high activation values are found near the red object, which is the furthest from the blue object. For other examples, see the Appendix.


We also checked RN if the proper object pairs are used to solve the question. However, it was possible to see that object pairs that do not have much significance in addressing the given question have high activation values of $g_{\theta}$. This indicates that there is lack of interpretability that can verify the reasoning is done soundly. See the Appendix for detailed examples.

\subsection{Robustness on image size and object sparsity}
\label{sec:robust}
We tested how robust SARN is to object size and image size with the same model parameters as in Section 3.2, which are shown in Table 4. By varying image size while fixing object size, we can evaluate how the model deals with sparsity. As image size gets bigger, more and more objects (pixels in the feature maps) will correspond to blank spots. When comparing the configurations where object size and image size are roughly in the same proportion, we can evaluate how the model deals with the granularity of object representation

We first tested robustness to sparsity by varying the image size to 64, 75, 128 while fixing the object size to 5. The baseline model and RN have similar performance on non-relational questions across all image sizes. However, they show worse results on relational questions as image size gets bigger. SARN is relatively robust and even shows higher accuracy with bigger image size.

Next, we tested robustness to granularity by changing the size of image and objects with the same proportion of image size-object size (75-5, 64-4, 128-8). Objects in the configuration (128-8) will be represented by more pixels than in (64-4). In case of RN, this will make each object (pixel) represent only a fraction of the original object in the image. However, SARN takes a soft-attention weighted representation for the reference object and is thus robust to how many pixels represent an original object in the image. The results reflect this: RN shows lower performance as the image size gets bigger. SARN shows stronger performance as the image size gets bigger, especially in relational questions.


\section{Conclusion}
We propose an attention module augmented relational network called SARN(Sequential Attention Relational Network) that implements an efficient sequential reasoning process of (1) finding the reference object and (2) extracting relevant relationships between the reference object and other objects. This greatly reduces the computational and memory requirements of \citep{santoro2017simple}, which computes all object pairs. It shows higher accuracy on the modified Sort-of-CLEVR dataset than other models, especially on relational questions. Also by inspecting the attention map and relational module, we can verify that the reasoning process is interpretable.

\clearpage
\bibliographystyle{plainnat} 
\bibliography{reference} 
\small
\clearpage

\section{Appendix}
\label{sec:appendix}

\subsection{test accuracy by question type}
\begin{table}[h]
\centering
\label{tbl:acc_non_rel}
\caption{Test accuracy: non-relational questions}
\begin{tabular}{@{}lllll@{}}
\toprule
model                                    & horizontal     & vertical       & shape          & non-rel        \\ 
\midrule
SARN & \textbf{99.92} & \textbf{99.67} & \textbf{99.92} & \textbf{99.84} \\ 
RN                      & \textbf{99.92} & \textbf{99.67} & 99.83          & 99.81          \\
base line                                & 96.33          & 96.58          & 99.83          & 97.58          \\ 
\bottomrule
\end{tabular}
\end{table}

\begin{table}[h]
\centering
\label{tbl:acc_rel}
\caption{Test accuracy: relational questions}
\begin{tabular}{@{}lllllll@{}}
\toprule
model                                    & cl\_col        & cl\_sh         & fur\_col       & fur\_sh        & count        & rel            \\ 
\midrule
SARN & \textbf{90.75} & \textbf{93.92} & \textbf{93.75} & \textbf{96.33} & 99.67        & \textbf{94.88} \\ 
RN                       & 86.33          & 88.42          & 84.17          & 90.25          & \textbf{100} & 89.83          \\ 
base line                                & 84.92          & 88.50          & 67.83          & 79.25          & 99.33        & 83.97          \\ 
\bottomrule
\end{tabular}
\end{table}

\subsection{Image resolution robustness}
\begin{table}[ht]
\label{tbl:robust}
\caption{image size-object size}
\centering
\begin{tabular}{@{}llllll@{}}
\toprule
         &         & 64\_4  & 128\_8 & 64\_5  & 128\_5 \\
         
         \midrule
SARN     & non-rel & 0.9970 & 0.9999 & 0.9948 & 0.9988 \\
         & rel     & 0.8949 & 0.9440 & 0.8370 & 0.8669 \\
         & total   & 0.9345 & 0.9650 & 0.8970 & 0.9163 \\
         \midrule
         
RN       & non-rel & 0.9944 & 0.9981 & 0.9964 & 0.9931 \\
         & rel     & 0.8415 & 0.8207 & 0.8430 & 0.7719 \\
         & total   & 0.8989 & 0.8872 & 0.9005 & 0.8555 \\
         \midrule
baseline & non-rel & 0.9941 & 0.9972 & 0.9933 & 0.9978 \\
         & rel     & 0.8120 & 0.8625 & 0.8163 & 0.8532 \\
         & total   & 0.8803 & 0.9130 & 0.8827 & 0.9074 \\
        \bottomrule
\end{tabular}
\end{table}

\clearpage
\subsection{Proposed model $a_{\psi}$ and $g_{\theta}$ channel sum plot}
\begin{figure*}[ht]
\centering
\begin{subfigure}[b]{\textwidth}
\centering   
\includegraphics[width=\textwidth]{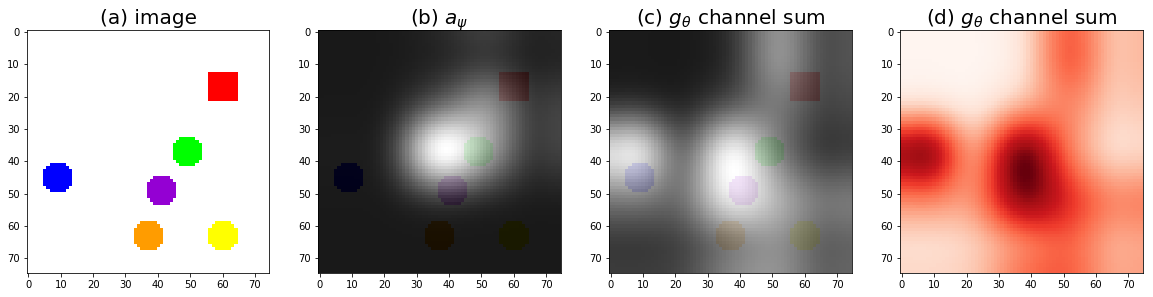}  
  \caption{What is shape of the object closest to the green object?}  
  \label{fig:attention_blue_furthest_shape}
\end{subfigure}

\begin{subfigure}[b]{\textwidth}
\centering   
\includegraphics[width=\textwidth]{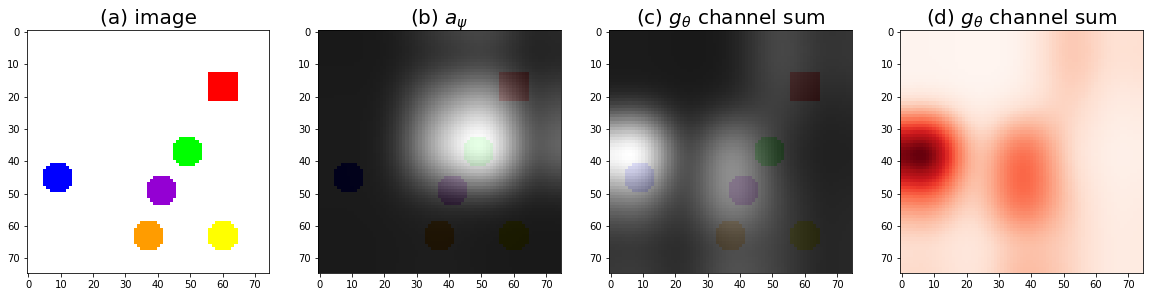}
  \caption{What is shape of the object furthest to the green object?}  
  \label{fig:attention_blue_furthest_shape}
\end{subfigure}

\begin{subfigure}[b]{\textwidth}
\centering   
\includegraphics[width=\textwidth]{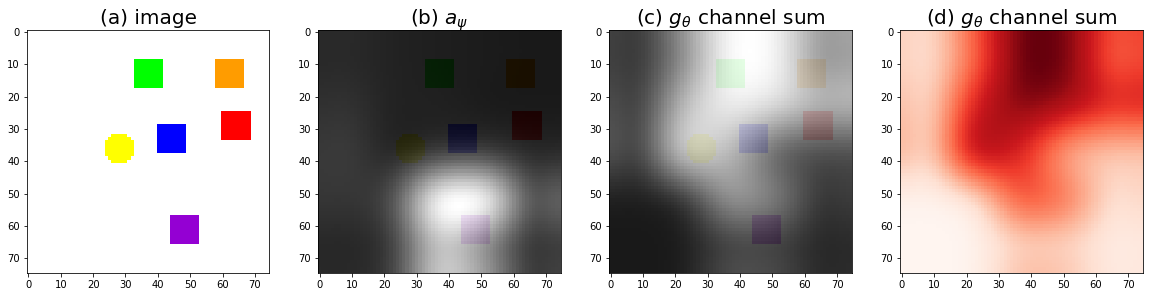}
  \caption{What is the number of objects that has the same shapoe as the violet object?}  
  \label{fig:attention_blue_furthest_shape}
\end{subfigure}

\begin{subfigure}[b]{\textwidth}
\centering   
\includegraphics[width=\textwidth]{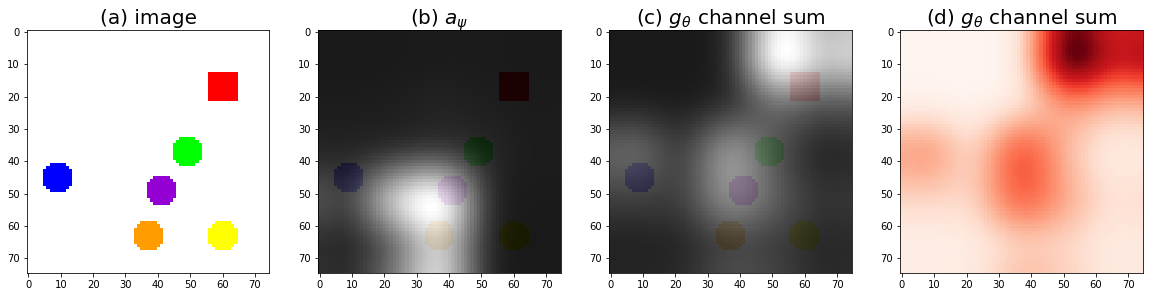}
  \caption{What is color of the object closest to the orange object?}  
  \label{fig:attention_orange_furthest_color}
 
\end{subfigure}
\caption{\textbf{Sample attention maps and relational module activation}: additional examples}
\label{fig:appendix_plots}
\end{figure*}

\begin{figure*}[ht]
\centering
\begin{subfigure}[b]{\textwidth}
\centering   
\includegraphics[width=\textwidth]{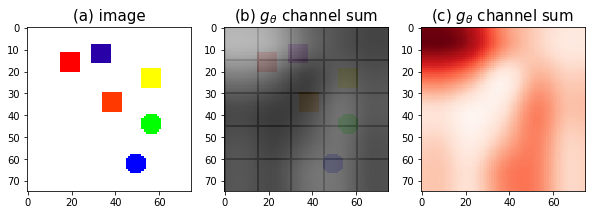}
  \caption{question: What is shape of the object that is furthest from the red object?}  
  \label{fig:attention_blue_furthest_shape}
\end{subfigure}

\begin{subfigure}[b]{\textwidth}
\centering   
\includegraphics[width=\textwidth]{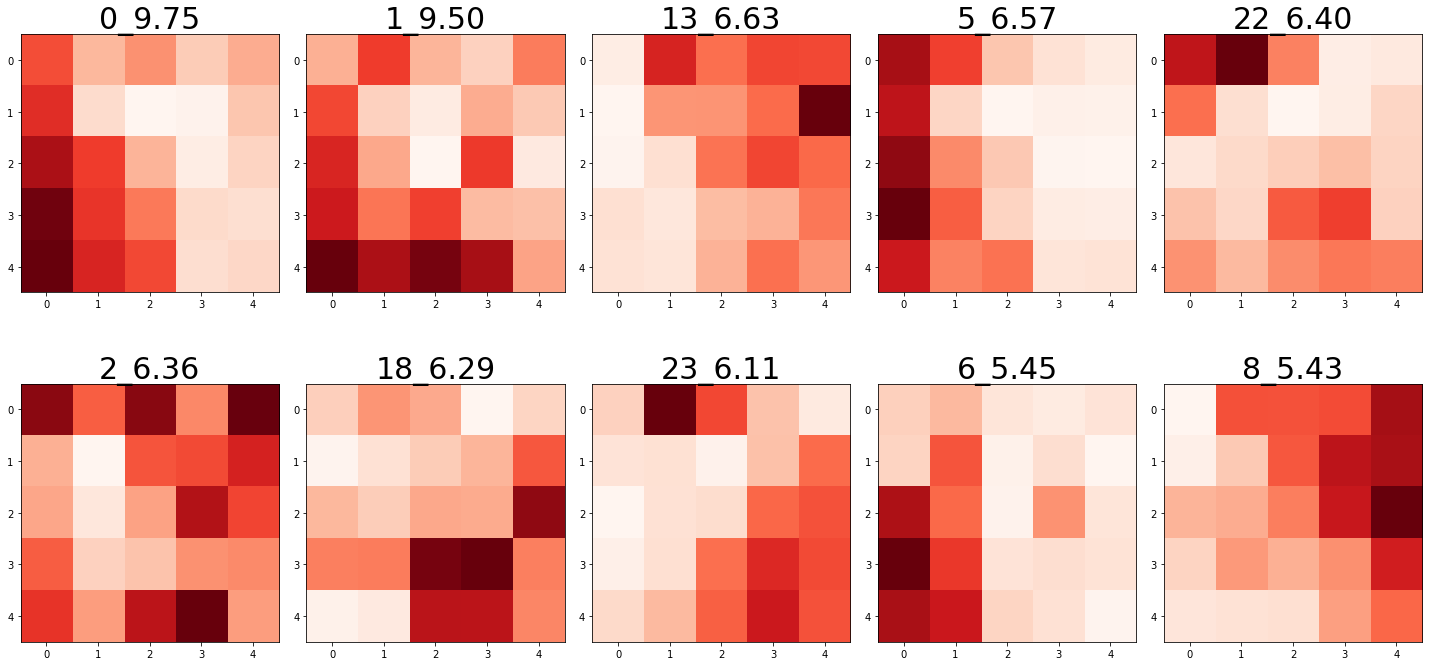} 
  \caption{$g_{\theta}$ plot for each object pair}
  \label{fig:attention_blue_furthest_shape}
\end{subfigure}
\caption{\textbf{Relational module activation of RN}: The three figures above show the $g_{\theta}$ output summed up across channels. The 8 figures below show the pairs with the highest activation values. The number above each figure shows the object that is paired with others. The first figure is the $g_{\theta}$ output of object pairs that is paired with the object 0. This tells that objects paired with 0 had the biggest summed up activation value of 9.75. However since the blue object is the furthest from the red object, object 23 should have been the relation that is most critical to solving the problem.}
\label{fig:rn_plots}
\end{figure*}




\end{document}